%% file: main.tex
\documentclass[10pt,twocolumn,letterpaper]{article}
\usepackage{multirow}   
\usepackage{algorithmic}
\usepackage{booktabs}   
\usepackage{array}      
\usepackage{booktabs}       
\usepackage[table]{xcolor}  
\usepackage{multirow}       
\usepackage{amsmath,amssymb,booktabs,multirow,graphicx,algorithm,algorithmic,xcolor}
\usepackage{comment}
\usepackage{graphicx}
\usepackage{arydshln}
\usepackage{tikz}
\usetikzlibrary{arrows.meta,positioning,fit,shapes.misc,shapes.geometric,calc}
\usepackage{colortbl}
\usepackage{amssymb}
\usepackage{algorithmic}
\usepackage{algorithm}
\usepackage{cuted}     
\usepackage{capt-of}
\usepackage[table]{xcolor}
\usepackage{booktabs}
\usepackage[table]{xcolor}
\usepackage{multirow}
\usepackage{siunitx}
\usepackage{booktabs}
\usepackage[table]{xcolor}
\usepackage{setspace}
\usepackage{lipsum}

\usepackage{cvpr}     


\definecolor{cvprblue}{rgb}{0.21,0.49,0.74}
\usepackage[pagebackref,breaklinks,colorlinks,allcolors=cvprblue]{hyperref}
\usepackage{comment}

\title{Vote-in-Context: Turning VLMs into Zero-Shot Rank Fusers}

\author{%
\makebox[\textwidth][c]{%
  \begin{tabular}{@{}c c c c c@{}}
  Mohamed Eltahir$^{1}$ &
  Ali Habibullah$^{1}$ &
  Lama Ayash$^{1,2}$ &
  Tanveer Hussain$^{3}$\footnotemark[1] &
  Naeemullah Khan$^{1}$\footnotemark[1] 
  \end{tabular}%
}%
\\[0.45em]
$^{1}$ King Abdullah University of Science and Technology (KAUST), Thuwal, Saudi Arabia \\
$^{2}$ Department of Computer Science, King Khalid University (KKU), Abha, Saudi Arabia \\
$^{3}$ Department of Computer Science, Edge Hill University, Ormskirk, England \\
\texttt{\{mohamed.hamid, ali.habibullah, lama.ayash\}@kaust.edu.sa} \\
\texttt{hussaint@edgehill.ac.uk, naeemullah.khan@kaust.edu.sa}}

\begin{document}

\twocolumn[{
  \renewcommand\twocolumn[1][]{#1}
  \maketitle
  \begin{center}
  \vspace{-2.2em}
    \captionsetup{type=figure}
    \includegraphics[width=0.90\textwidth]{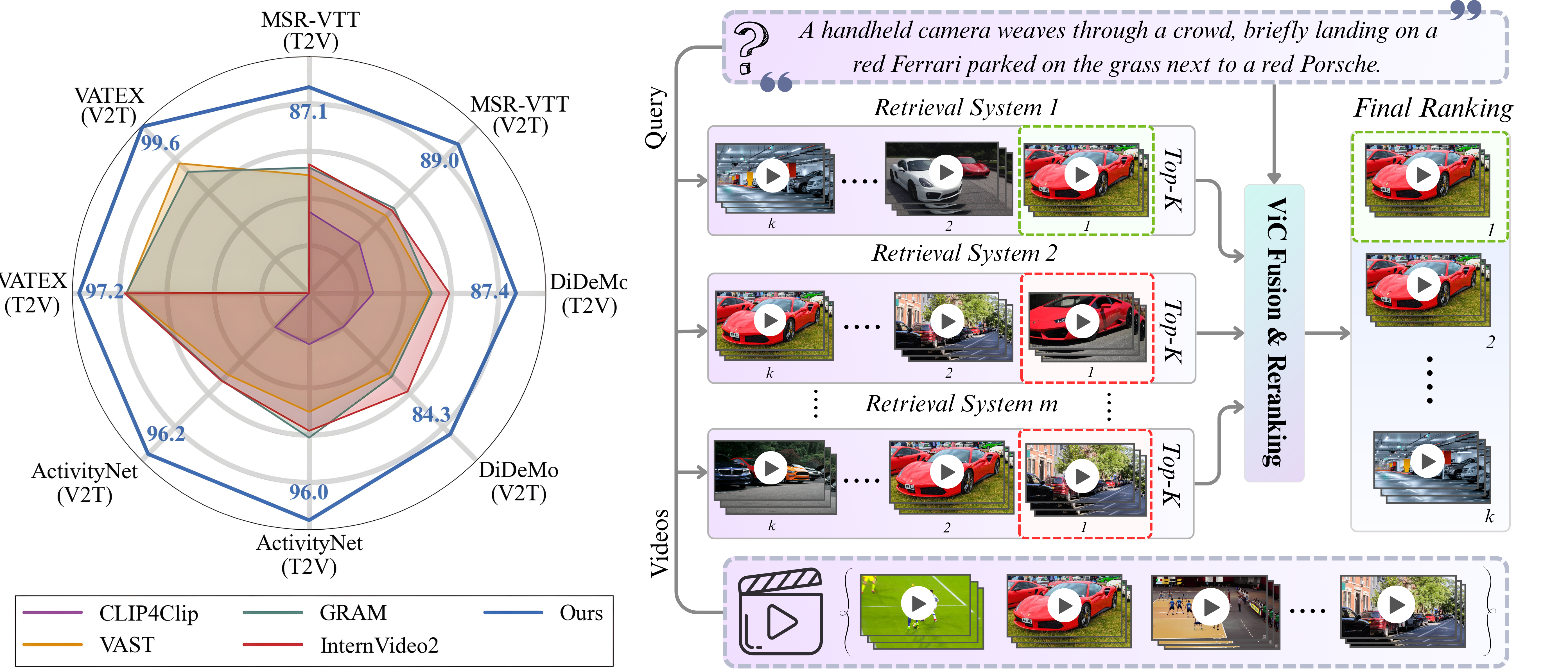}
    \captionof{figure}{Left: R@1 for T2V/V2T on MSR-VTT, DiDeMo, VATEX, and ActivityNet versus strong baselines. Right: Qualitative example where multi-retriever outputs are fused and re-ranked (ViC) to obtain the final list.}
    \label{fig:radar}
  \end{center}
}]

\newcommand\blfootnote[1]{%
  \begingroup
  \renewcommand\thefootnote{}\footnote{#1}%
  \addtocounter{footnote}{-1}%
  \endgroup
}

\input{0_abstract_new}

\blfootnote{* Principal Investigator (PI)}

\input{1_intro_new}

\makeatletter
\def\thebibliography#1{\section*{\refname\@mkboth{\refname}{\refname}}\small
  \def\list@biblabel##1{##1.}%
  \list{\@biblabel{\arabic{enumiv}}}%
       {\settowidth\labelwidth{\@biblabel{#1}}%
        \leftmargin\labelwidth
        \advance\leftmargin\labelsep
        \@openbib@code
        \usecounter{enumiv}%
        \let\p@enumiv\@empty
        \renewcommand\theenumiv{\arabic{enumiv}}}%
  \sloppy\clubpenalty4000\widowpenalty4000%
  \sfcode`\.\@m} 
\makeatother

\input{2_methodology}

\input{3_results}

\end{document}

%% file: 0_abstract_new.tex
\begin{abstract}
In the retrieval domain, candidates' fusion from heterogeneous retrievers is a long-standing challenge, particularly for complex, multi-modal data such as videos. While typical fusion techniques are training-free, they rely solely on rank or score signals, disregarding candidates' representations. This work introduces \emph{Vote-in-Context (ViC)}, a generalized, training-free framework that re-thinks list-wise reranking and fusion as a zero-shot reasoning task for a Vision-Language Model (VLM). The core insight is to serialize both \emph{content evidence} and \emph{retriever metadata} directly within the VLM's prompt, allowing the model to adaptively weigh retriever consensus against visual-linguistic content. We demonstrate the generality of this framework by applying it to the challenging domain of cross-modal video retrieval. To this end, we introduce the \emph{S-Grid}, a compact serialization map that represents each video as an image grid, optionally paired with subtitles to enable list-wise reasoning over video candidates. ViC is evaluated both as a \emph{single-list reranker}, where it dramatically improves the precision of individual retrievers, and as an \emph{ensemble fuser}, where it consistently outperforms strong baselines like CombSUM. Across video retrieval benchmarks including ActivityNet and VATEX, the framework establishes new state-of-the-art zero-shot retrieval performance, demonstrating its effectiveness in handling complex visual and temporal signals alongside text. In zero-shot settings, ViC achieves Recall@1 scores of \emph{87.1\%} (t2v) / \emph{89.0\%} (v2t) on MSR-VTT and \emph{99.6\%} (v2t) on VATEX, representing massive gains of up to +40 Recall@1 over previous state-of-the-art baselines. We present ViC as a simple, reproducible, and highly effective recipe for turning modern VLMs into powerful zero-shot rerankers and fusers. Code and resources are publicly available at: \url{https://github.com/mohammad2012191/ViC}
\end{abstract}

%% file: 1_intro_new.tex
\section{Introduction}
\label{sec:intro}
The digital age is characterized by an exponential growth in complex data. This includes vast repositories of unstructured text, which are central to modern applications like Retrieval-Augmented Generation (RAG)~\cite{lewis2020retrieval}, as well as complex multimodal data, such as video, which integrates visual, auditory, and temporal signals~\cite{nguyen-etal-2024-video}. The sheer volume and variety of this data make efficient organization, retrieval, and analysis increasingly difficult~\cite{zhu2023deeplearning}.

To address this, modern retrieval systems seek to align natural language queries with semantically relevant content, enabling users to efficiently locate desired material within large-scale data repositories. Despite significant progress, this task remains challenging due to the complexity of the data itself, such as high dimensionality or temporal structure, and the queries, such as sparsity or ambiguity.

Considering this complexity, a two-stage retrieval paradigm is commonly adopted~\cite{nogueira2019passage}. In the first stage, a computationally efficient retriever, such as a dual-encoder, retrieves a broad pool of candidate items. In the second stage, a more powerful yet computationally expensive re-ranker refines this shortlist to enhance precision. This two-stage pipeline has become a standard framework in both text retrieval and retrieval-augmented generation~\cite{qin2023large,chen2017reading,lewis2020retrieval}. Furthermore, two-stage pipelines enable using multiple diverse retrievers as a first stage. Fusing their results based on ranks or scores, using Reciprocal Rank Fusion (RRF)~\cite{10.1145/1571941.1572114} or CombSUM/CombMNZ~\cite{fox1994combination}, respectively, is a common technique that typically offers significant performance gains~\cite{e25010132}.

However, applying this two-stage template to complex, multimodal data is non-trivial, revealing limitations in the second stage. First-stage retrievers, while computationally efficient, typically rely on global embeddings and may rank irrelevant candidates highly because they fail to capture or verify all query-specific details. A second stage is therefore essential, but it presents two key challenges. First, conventional rerankers for a single list are often costly, require fine-tuning on in-domain data, or are tied to a specific retriever's features~\cite{tian2024towards}. Second, when ensembling multiple retrievers, conventional fusion methods are "content-blind," as they operate only on rank/score signals while ignoring the candidates' rich content. These limitations motivate the need for a universal, training-free framework capable of acting as both a content-aware reranker and fuser.

Recent advances in large-scale, instruction-following Language models offer a promising solution. In text retrieval, Large Language Models (LLMs) have proven to be powerful zero-shot listwise rerankers, as seen in work like RankGPT~\cite{qin2023large}. This paradigm extends to Vision-Language Models (VLMs), such as InternVL 3.5~\cite{wang2025internvl3} and Qwen-VL~\cite{bai2023qwen}, which demonstrate strong zero-shot reasoning and cross-modal alignment capabilities. By adapting videos into a format interpretable by VLMs, these models can themselves serve as powerful zero-shot relevance estimators.

To this end, we introduce \textbf{Vote-in-Context (ViC)}, a generalized, training-free framework that utilizes a frozen VLM as a universal, list-wise reranker and fuser. Instead of collapsing $M$ ranked lists with a fixed formula, such as Reciprocal Rank Fusion (RRF), \textbf{ViC} serializes both \emph{content evidence} (such as images, text) and \emph{retriever metadata} (such as, per-list ranks, cross-list multiplicity) directly into the VLM's prompt, allowing it to adaptively weigh all signals.

In this paper, we apply \textbf{ViC} to video retrieval. We propose the \textbf{S-Grid}, a compact content serialization map that represents a video as a single image grid of uniformly sampled frames, optionally paired with subtitles. This S-Grid acts as the VLM-readable \emph{content evidence} for each video candidate.

The framework operates in two modes. First, as a powerful single-list reranker ($M=1$), where \textbf{ViC} uses S-Grids to re-evaluate the top-$K$ items from one retriever. Second, as a novel ensemble fuser ($M>1$), where \textbf{ViC} constructs a candidate list by interleaving multiple retrievers. This assembly explicitly encodes rank and consensus metadata in the list order and item multiplicity, allowing the VLM to weigh these signals jointly with the S-Grid content evidence. The experiments show this combination yields massive gains, saturating several benchmarks in a zero-shot settings.

The main contributions of this work are summarized as follows:
\begin{itemize}
    \item We propose \textbf{Vote-in-Context (ViC)}, a generalized, training-free framework that turns a frozen VLM into a powerful list-wise reranker and fuser by serializing both content and retriever metadata into its prompt.
    \item We introduce the \textbf{S-Grid}, a compact and effective video representation that serves as the content serialization map for \textbf{ViC}, enabling VLM-based reasoning over video without costly sequence processing.
    \item We comprehensively evaluate \textbf{ViC} in both its $M=1$ (single-list) and $M>1$ (fusion) modes. We show that \textbf{ViC} as a reranker ($M=1$) dramatically improves all single backbones, and \textbf{ViC} as a fuser ($M>1$) consistently outperforms strong baselines like RRF and CombSUM.
    \item We release our framework and evaluation protocols, including an extensive analysis of \textbf{ViC}'s scaling properties, its sensitivity to context size, and the performance of different assembly strategies.
\end{itemize}

This paper is organized as follows. Section~\ref{sec:related_work} reviews related work. Section~\ref{sec:method} details the proposed \textbf{ViC} framework and its application for video retrieval. Section~\ref{sec:results} presents the experimental results and ablation studies, followed by a discussion of the framework-s limitations and future directions.

\begin{figure*}
    \centering
    \includegraphics[width=1\linewidth]{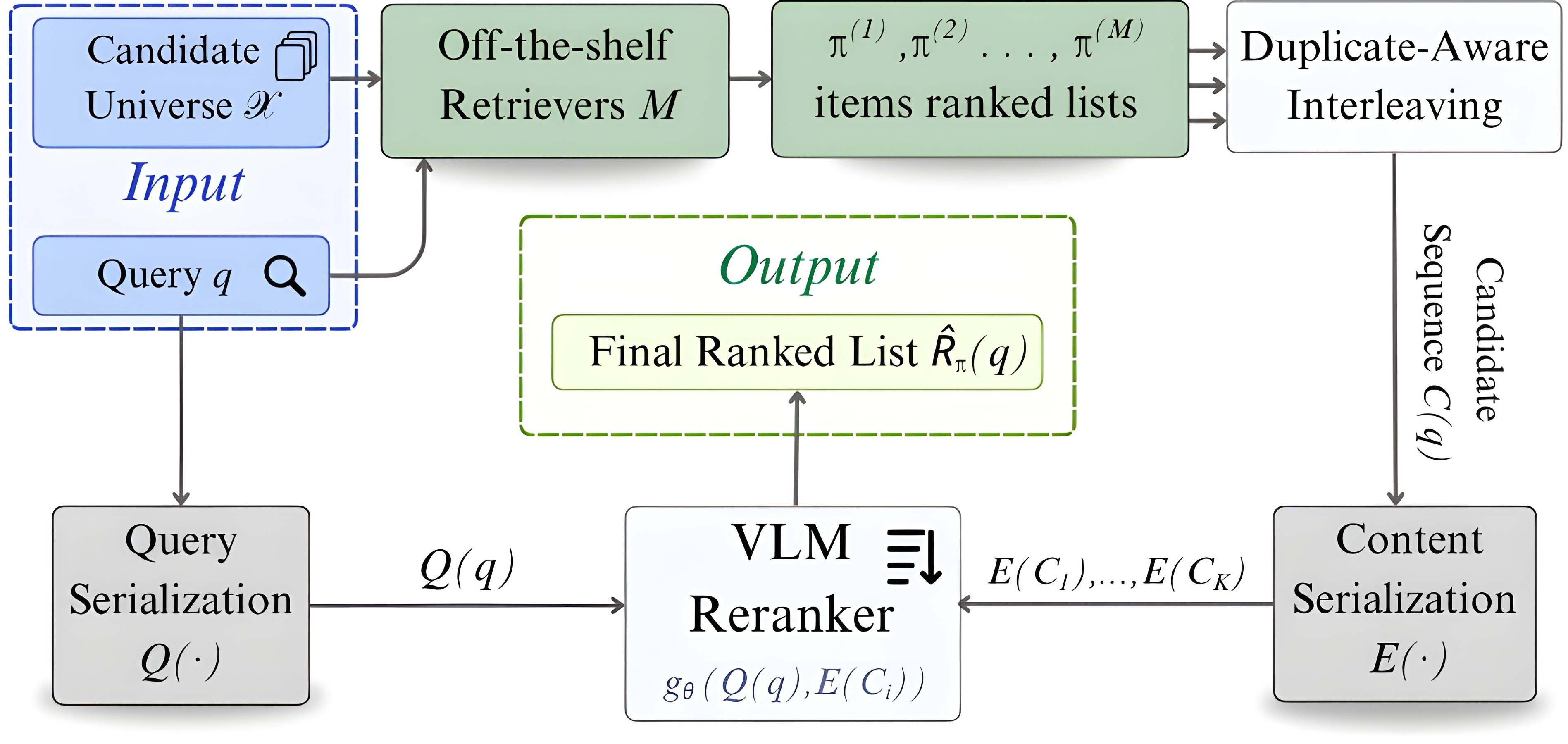}
    \caption{The \textbf{Vote-in-Context (ViC)} framework. A \texttt{VLM Reranker} jointly weighs serialized content ($Q(\cdot)$, $E(\cdot)$) against retriever metadata (rank, multiplicity) encoded in the \texttt{Candidate Sequence $C(q)$} by \texttt{Duplicate-Aware Interleaving} step to produce the final ranking $\widehat{R}(q)$.}
\label{fig:vic_framework}
\end{figure*}

\section{Related Work}
\label{sec:related_work}

Modern video retrieval has evolved from early architectures that coupled temporal attention mechanisms with language encoders~\cite{yu2018joint,zhu2020actbert} to large-scale unified pretraining~\cite{bain2021frozen,xu2021videoclip}. CLIP-style adaptations, which transfer powerful image-text encoders to video, such as CLIP4Clip~\cite{luo2021clip4clip} and X-CLIP~\cite{ma2022x}, became a dominant paradigm for zero-shot retrieval. Recent foundation-scale systems have pushed recall even further by incorporating broader multi-modality, such as audio/subtitles in VAST~\cite{chen2023vast}, and larger, video-specific backbones, such as InternVideo2~\cite{wang2024internvideo2}. These models serve as the "first-stage" retrievers in our work. However, they primarily rely on matching coarse, global representations. While this is computationally efficient for rapidly narrowing a large search space to a high-recall candidate set, this reliance on coarse similarity means they can struggle to capture fine-grained, query-specific details, often leading to imprecise top rankings. 

Building upon these first-stage retrievers, subsequent research has explored two-stage architectures that refine coarse candidate sets. When multiple first-stage lists are available, they must be fused. Classical fusion methods operate at the score level, such as CombSUM/CombMNZ~\cite{fox1994combination} or the rank level, such as Reciprocal Rank Fusion (RRF)~\cite{10.1145/1571941.1572114}. These methods are simple, robust, and widely used baselines for aggregating ranked lists, making them key points of comparison for our fusion method. However, despite their efficiency, such methods assume a fixed weighting formula and hyperparameters, such as RRF-s $k$, and operate solely on rank or score signals, leaving other modalities unexploited.

The emergence of large language models (LLMs) and Vision-Language Models (VLMs) has introduced a new paradigm for re-ranking in retrieval systems. LLMs have recently demonstrated strong zero-shot, list-wise re-ranking capabilities in text retrieval, achieving substantial performance gains by reasoning jointly over ranked lists of passages~\cite{qin2023large,zhang2025rank,adeyemi2023zero}. Adapting this paradigm to video, however, is non-trivial as VLMs cannot process raw videos. To overcome this, several studies have shown that representing a video clip as a grid of sampled frames enables image-centric VLMs to reason effectively about temporal dynamics~\cite{kim2024image}. At the same time, modern instruction-following VLMs, such as InternVL~\cite{wang2025internvl3} and Qwen-VL~\cite{bai2023qwen}, provide the robust zero-shot, multimodal alignment required to make such designs practical.

%% file: 2_methodology.tex
\begin{figure*}
    \centering
    \includegraphics[width=1\linewidth]{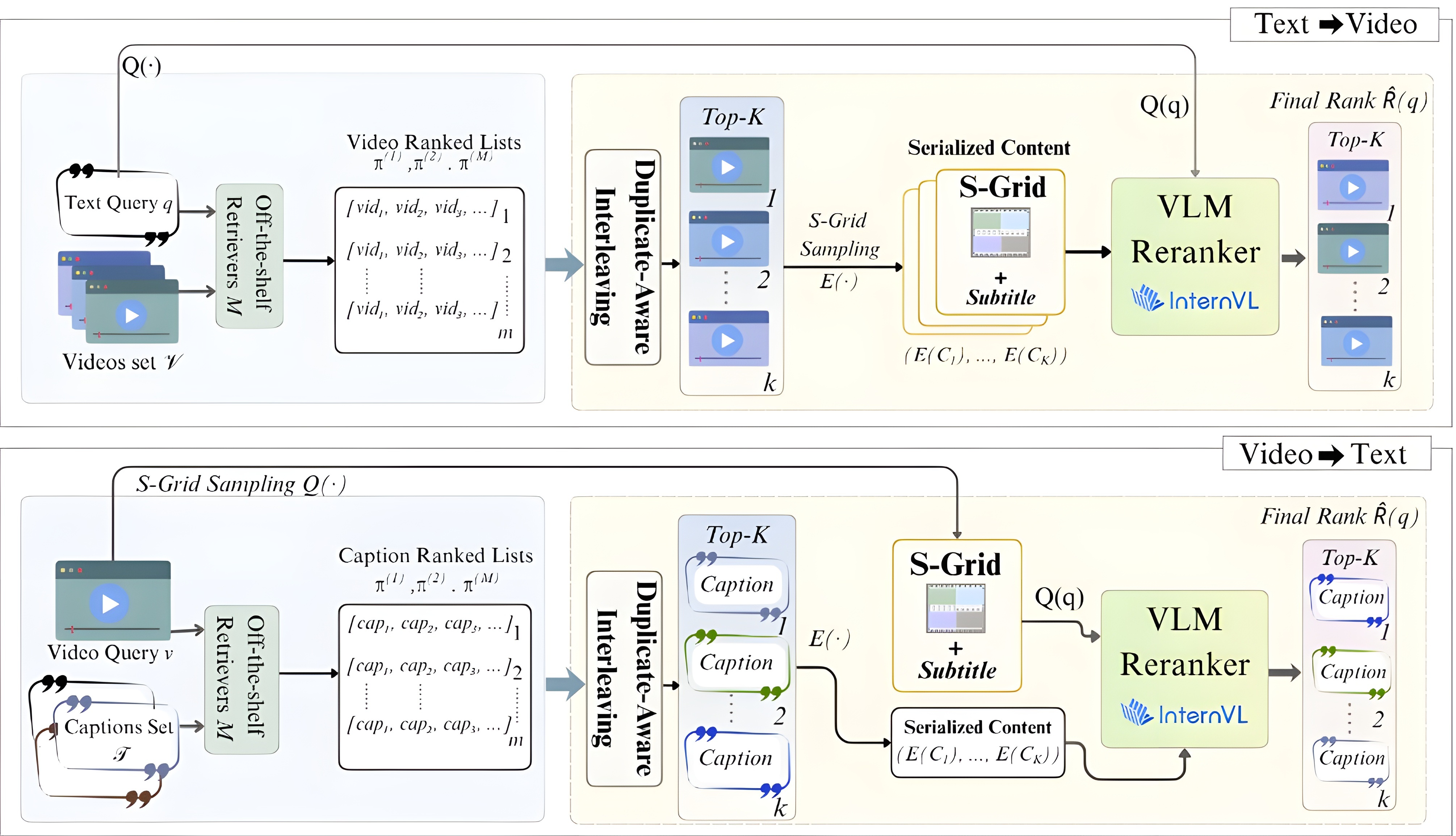}
    \caption{The \textbf{Vote-in-Context (ViC)} framework applied for Text-to-Video (t2v, top) and Video-to-Text (v2t, bottom). The left block shows the initial retrieval stage. The right block (green) shows our \textbf{ViC} framework. The serialization maps ($Q(\cdot)$, $E(\cdot)$) are modality-dependent: \texttt{S-Grid Sampling} is applied to video inputs, while text inputs use the identity.}
\label{fig:vid_ret_pipeline}
\end{figure*}

\section{Methodology}
\label{sec:method}
This paper introduces \textbf{Vote-in-Context (ViC)}, a general, training-free, and multimodal framework that utilizes the VLM reasoning capabilities discussed in \S\ref{sec:related_work} to solve the ranked-list fusion problem. Rather than collapsing lists with a fixed formula, \textbf{ViC} provides a uniform candidate prompt to the VLM containing both: (a) \emph{content evidence} (such as images/text), and (b) \emph{retriever metadata}, including ranks and cross-list multiplicity encoded directly in the prompt. This approach stands in contrast to classical, non-content-aware fusion methods (such as RRF or CombSUM), which operate only on rank/score signals and ignore candidate content.

The VLM receives this meta-signal alongside the candidates' content and implicitly weighs retriever metadata versus content evidence on a per-query basis in a zero-shot setting. A candidate's rank is conveyed by each list's order, while cross-list consensus is represented by allowing duplicates to appear in the candidate set. Compared to the traditional fusion methods, \textbf{ViC} is hyperparameter-free and modality-aware, yielding per-query decisions that adaptively weight all available signals. The idea is modality-agnostic, requiring only that candidates can be serialized into a VLM-readable prompt (such as passages for text search, images with metadata, tables, or audio transcripts).

To demonstrate this framework's generality, \textbf{ViC} is applied to video retrieval as a second-stage fuser and re-ranker. The framework fuses candidate results from multiple first-stage retrievers and serializes each video into a uniform visual-linguistic representation, termed the S-Grid. The VLM is subsequently employed to produce a list-wise permutation of the candidate set. Both text-to-video (t2v) and video-to-text (v2t) retrieval tasks are evaluated within a two-stage pipeline consisting of dual-encoder recall followed by ViC-based re-ranking.


\subsection{Problem Setup and Notation}
\label{ssec:problem_setup}

The \textbf{ViC} fusion framework is formalized as follows. Let $\mathcal{X}$ denote the universe of candidate items (such as videos or text passages). For a given query $q$, assume access to $M$ retrievers, $\mathcal{M} = \{1, \dots, M\}$. Each retriever $m \in \mathcal{M}$ returns a ranked list of items drawn from $\mathcal{X}$:
$$
L_m(q) = \big(x_{m,1}, x_{m,2}, \dots, x_{m,n_m}\big), \quad \text{where } x_{m,j} \in \mathcal{X}.
$$
\textbf{ViC} aggregates the $M$ ranked lists into a single fused ranking of target length $K$.

\paragraph{Candidate Assembly and Metadata Encoding.}
The process begins by constructing a single \emph{candidate sequence} $C(q)$ of length $K$.
This sequence retains both the rank and multiplicity metadata from the initial retrieval lists.
Define a per-list depth as $k_{\max} = \lceil K / M \rceil$, and truncate each list accordingly before assembling the final candidate sequence.
$$
\mathrm{Top}_{k_{\max}}(L_m) = \big(x_{m,1}, \dots, x_{m,\min(k_{\max}, n_m)}\big).
$$
The candidate sequence $C(q)$ is formed by a round-robin (RR) interleaving of these truncated lists, preserving duplicates:
$$
C(q) = \mathrm{RR}_K\!\big(\,\mathrm{Top}_{k_{\max}}(L_1), \dots, \mathrm{Top}_{k_{\max}}(L_M)\big).
$$
The $\mathrm{RR}_K(\cdot)$ operator appends items in the order $\big(x_{1,1}, x_{2,1}, \dots, x_{M,1}, x_{1,2}, \dots\big)$, skipping any exhausted lists, and truncates the final sequence to length $K$. This sequence $C(q) = (C_1, \dots, C_K)$ inherently encodes retriever metadata: per-list rank is signaled by position, and cross-list consensus is signaled by an item's multiplicity, $\mu_C(x) = \sum_{i=1}^{K} \mathbf{1}\{C_i = x\}$.

\paragraph{VLM Re-ranking.} The sequence is passed to a frozen, list-wise VLM $g_{\Theta}$ for reranking. Let $E(\cdot)$ be the \emph{content serialization map} that converts an item $x \in \mathcal{X}$ into its VLM-readable format (i.e., the content evidence), and let $Q(q)$ be the serialized query. The VLM computes a permutation $\hat{\pi} \in \mathfrak{S}_K$, where $\mathfrak{S}_K$ is the set of all permutations of the indices $\{1, \dots, K\}$:
$$
\hat{\pi} = g_{\Theta}\!\Big(Q(q), \; \big(E(C_1), E(C_2), \dots, E(C_K)\big)\Big).
$$
The final fused and reranked output $\widehat{R}(q)$ is the sequence $C$ reordered by this permutation:
$$
\widehat{R}(q) = \big(C_{\hat{\pi}(1)}, \, C_{\hat{\pi}(2)}, \, \dots, \, C_{\hat{\pi}(K)}\big).
$$
See Fig.~\ref{fig:vic_framework} for a high-level overview.

\paragraph{Special Case: Single-List Reranking ($M=1$).}
The \textbf{ViC} framework naturally handles the standard single-list reranking task as a special case. When $M=1$, the round-robin interleaving simplifies, and the candidate sequence $C(q)$ becomes the standard top-$K$ list from the single retriever:
$$
C(q) = \mathrm{Top}_K(L_1(q)) = (x_{1,1}, \dots, x_{1,K}).
$$
The VLM call and final output $\widehat{R}(q)$ remain identical. In this $M=1$ setting, \textbf{ViC} functions as a pure list-wise reranker. The VLM's decision is based solely on the \emph{content evidence} $E(\cdot)$ of the candidates relative to the query, as the retriever metadata signals (cross-list multiplicity and rank-of-ranks) are absent.

\subsection{Applying ViC to Video Retrieval.}
Applying \textbf{ViC} to video retrieval requires a method to serialize video candidates into a VLM-readable format. This section first defines this video representation, the S-Grid, and then maps it to the \textbf{ViC} framework.

\begin{figure}
    \centering
\includegraphics[width=1\linewidth]{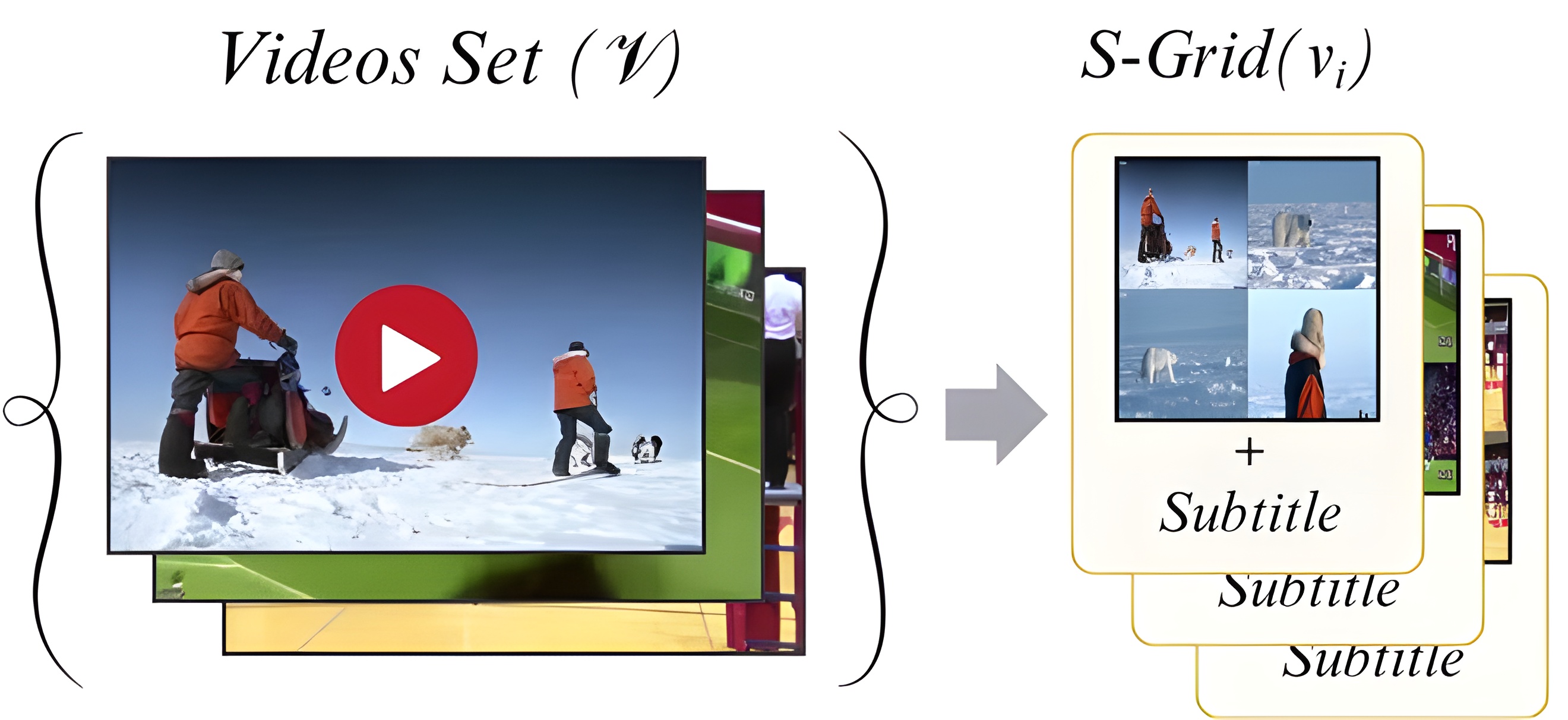}
    \caption{The S-Grid representation.}
\label{fig:sgrid}
\end{figure}

\paragraph{S-Grid: A Uniform Video Prompt.} A video $v$ is represented as a regular grid of uniformly sampled frames composited into a single $H\times W$ image, optionally paired with a subtitle or Automated Speech Recognition (ASR) string $a_v$ (if available). Let $s$ denote the grid dimension (i.e., the grid has $s \times s$ cells). Given video length $F$ frames, $s^2$ frame indices $\{t_i\}_{i=1}^{s^2}$ are selected uniformly via $t_i=\lfloor (i-1)\tfrac{F}{s^2-1}\rfloor$. These frames are extracted, resized to $\lfloor H/s \rfloor \times \lfloor W/s \rfloor$, and tiled in row-major order to form an $H \times W$ canvas, denoted as $\mathrm{Grid}(v; s)$.
When a subtitle $a_v$ is available, it is concatenated to the textual prompt as an auxiliary input. This representation, visualized in Fig.~\ref{fig:sgrid}, is denoted as:
\[
\mathrm{S\text{-}Grid}(v) \;=\; \big(\mathrm{Grid}(v; s), \; a_v \big),
\]
This design provides the VLM with both visual snapshots and audio transcripts within a single prompt.
Such a uniform interface enables a single VLM to process candidates retrieved from \emph{any} upstream model.

\paragraph{Formalizing the Video Retrieval Tasks.} The \textbf{ViC} framework is applied to cross-modal video retrieval, where the candidate universe consists of videos $\mathcal{V}$ and text captions $\mathcal{T}$. In the t2v retrieval task, the query $q \in \mathcal{T}$ is text ($Q(q) = q$), and the candidates $C_i \in \mathcal{V}$ are videos, which are serialized as follows:
$$
E(v) = \big(\mathrm{S\text{-}Grid}(v), a_v\big).
$$
In the v2t retrieval task, the query $q \in \mathcal{V}$ is a video serialized as $Q(q) = \big(\mathrm{S\text{-}Grid}(v), a_v\big)$, and the candidates $C_i \in \mathcal{T}$ are text captions, so the content map is the identity ($E(t) = t$). This bidirectional retrieval process is illustrated in Fig.~\ref{fig:vid_ret_pipeline}.

\paragraph{Cost.} The re-ranker processes one image per video candidate and a short text block per item. The complexity per query is $\mathcal{O}(K \cdot C_{\text{VLM}})$ where $K$ is the number of candidates, and $C_{\text{VLM}}$ is one forward pass cost. This cost is independent of the raw video length, as each video is represented by a single image, keeping the per-candidate cost effectively constant.
The approach is significantly lighter than frame-level cross-attention and permits larger candidate sets to be evaluated within the VLM's context window.

\subsection{List Fusion Strategies}
\label{ssec:fusion}

Given $M$ off-the-shelf retrievers that produce ranked lists for a query, two standard list-fusion baselines are examined and compared against the proposed \textbf{ViC}.

\paragraph{(a) Soft Voting (score fusion).}
When calibrated similarity matrices are available, normalize each score distribution per query using min-max scaling and aggregate the results with nonnegative weights:
\begin{align*}
\tilde{S}(q,\cdot) & \;=\; \sum_{m=1}^M w_m\,\mathrm{norm}\!\big(S^{(m)}(q,\cdot)\big), \\
\mathcal{C} & \;=\; \mathrm{TopK}\!\big(\tilde{S}(q,\cdot)\big).
\end{align*}
This family includes classical CombSUM/CombMNZ-style score fusion and serves as a strong yet simple baseline when scores are comparable across retrieval systems.

\paragraph{(b) Reciprocal Rank Fusion (RRF).}
When only heterogeneous \emph{ranked lists} are available, RRF assigns each item $x$ a fused score as 
\[
\mathrm{RRF}(x) \;=\; \sum_{m=1}^{M} \frac{1}{k + \mathrm{rank}_m(x)},
\]
with a small smoothing constant $k$ (commonly $k{=}60$), then returns the Top-$K$ unique items. 
\paragraph{(c) Ours: Vote-in-Context (ViC).}
As formally defined in \S\ref{ssec:problem_setup}, the \textbf{ViC} framework defers the fusion logic to the VLM itself.
Rather than collapsing ranked lists into a single aggregated score, as in Soft Voting or RRF, \textbf{ViC} serializes both the \emph{content evidence} $E(\cdot)$ and the \emph{retriever metadata} (rank, multiplicity) directly into the VLM prompt.
This design allows the frozen VLM to adaptively weigh all available signals on a per-query basis, thereby functioning as a training-free, multimodal fusion model.

The serialization process also provides practical control mechanisms.
A round-robin assembly based on $k_{\max}$ ensures balanced coverage across all retrievers, while the candidate sequence $C(q)$ can be optionally reordered to bias the VLM's early context, by prioritizing items from stronger backbones, for instance.
Such flexibility is inherently absent from fixed-formula fusion methods.

%% file: 3_results.tex
\begin{table*}[t]
\centering
\small
\setlength{\tabcolsep}{8pt}
\begin{tabular}{@{}ll cccccccc@{}}
\toprule
\multirow{2}{*}{\textbf{Backbone}} & \multirow{2}{*}{\textbf{Reranker Input}} &
\multicolumn{2}{c}{\textbf{MSR-VTT}} &
\multicolumn{2}{c}{\textbf{DiDeMo}} &
\multicolumn{2}{c}{\textbf{ActivityNet}} &
\multicolumn{2}{c}{\textbf{VATEX}} \\
\cmidrule(lr){3-4}\cmidrule(lr){5-6}\cmidrule(lr){7-8}\cmidrule(lr){9-10}
 & & t2v & v2t & t2v & v2t & t2v & v2t & t2v & v2t \\
\midrule
\multicolumn{10}{@{}l@{}}{\cellcolor{gray!15}\textsc{\textbf{Baselines (No Reranking)}}} \\
\addlinespace[1pt]
\rowcolor{gray!8}
CLIP4Clip & None & 34.4 & 29.9 & 27.1 & 20.3 & 21.6 & 20.3 & -- & -- \\ 
\rowcolor{gray!8}
VAST & None & 49.9 & 46.2 & 51.0 & 47.8 & 50.2 & 48.7 & 77.0 & 77.6 \\
\rowcolor{gray!8}
GRAM & None & 53.1 & 50.8 & 51.8 & 49.6 & 61.1 & 52.1 & 77.3 & 72.5 \\
\rowcolor{gray!8}
InternVideo2-6B & None & 54.5 & 49.5 & 59.2 & 58.8 & 58.2 & 52.4 & 80.7 & -- \\

\midrule
\multicolumn{10}{@{}l@{}}{\cellcolor{blue!15}\textsc{\textbf{With ViC Single-List Reranking ($M=1$)}} \textcolor{gray}{\small (InternVL 3.5 38B, Grid Size 3x3)}} \\
\addlinespace[1pt]
\multirow{2}{*}{CLIP4Clip} 
 & Grid   & 62.8 & 61.3 & 60.4 & 53.8 & 64.6 & 62.8 & -- & -- \\
 & S-Grid & 64.2 & 62.5 & -- & -- & -- & -- & -- & -- \\
\addlinespace[2pt]
\multirow{2}{*}{VAST} 
 & Grid   & 67.3 & 62.2 & 70.2 & 63.4 & 79.7 & 75.2 & 91.9 & 99.4 \\
 & S-Grid & 68.7 & 63.1 & -- & -- & -- & -- & 92.4 & \textbf{99.6} \\
\addlinespace[2pt]
\multirow{2}{*}{GRAM} 
 & Grid   & 75.4 & 72.3 & 70.9 & 63.9 & 82.4 & 77.2 & -- & -- \\
 & S-Grid & \textbf{76.2} & 73.6 & -- & -- & -- & -- & -- & -- \\
\addlinespace[2pt]
\multirow{2}{*}{InternVideo2-6B} 
 & Grid   & 74.0 & 74.1 & \textbf{78.1} & \textbf{70.7} & \textbf{89.8} & \textbf{84.9} & 95.5 & -- \\
 & S-Grid & 75.9 & \textbf{76.6} & -- & -- & -- & -- & \textbf{95.8} & -- \\
\bottomrule
\end{tabular}
\caption{Zero-shot t2v and v2t retrieval: R@1 for single backbones without reranking vs. with \textbf{ViC} as a single-list reranker ($M=1$). Bold indicates the best result for each benchmark.}
\label{tab:retrieval}
\end{table*}

\begin{table*}[t]
\centering
\small
\setlength{\tabcolsep}{8pt}
\begin{tabular}{@{}lcccccccc@{}}
\toprule
\multirow{2}{*}{\textbf{Method}} &
\multicolumn{2}{c}{\textbf{MSR-VTT}} &
\multicolumn{2}{c}{\textbf{DiDeMo}} &
\multicolumn{2}{c}{\textbf{ActivityNet}} &
\multicolumn{2}{c}{\textbf{VATEX}} \\
\cmidrule(lr){2-3}\cmidrule(lr){4-5}\cmidrule(lr){6-7}\cmidrule(lr){8-9}
& t2v & v2t & t2v & v2t & t2v & v2t & t2v & v2t \\
\midrule
\multicolumn{9}{@{}l@{}}{\cellcolor{gray!15}\textsc{\textbf{Baseline \& Traditional Fusion Methods}}} \\
\addlinespace[1pt]
\rowcolor{gray!8}
InternVideo2 (Prev. SOTA) & 54.5 & 49.5 & 59.2 & 58.8 & 58.2 & 52.4 & 80.7 & -- \\
\addlinespace[2pt]
RRF & 78.3 & 80.2 & 72.8 & 73.2 & \textbf{96.8} & \textbf{97.4} & 94.7 & -- \\
CombSUM & 84.4 & 83.0 & 80.4 & 83.1 & 95.8 & 95.2 & 96.1 & -- \\
CombMNZ & 85.3 & 86.9 & 78.0 & 80.8 & 95.0 & 92.2 & 96.4 & -- \\
\midrule
\multicolumn{9}{@{}l@{}}{\cellcolor{blue!15}\textsc{\textbf{Our VLM-based Reranking Method}}} \\
\addlinespace[1pt]
\textbf{ViC (No Duplicates)} & 84.2 & 80.7 & 85.5 & 76.1 & 94.8 & 91.9 & 96.1 & -- \\
\textbf{ViC} & \textbf{87.1} & \textbf{88.1} & \textbf{87.4} & \textbf{84.3} & 96.0 & 96.2 & \textbf{97.5} & -- \\
\bottomrule
\end{tabular}
\caption{Zero-shot t2v and v2t retrieval with ensemble fusion methods. All metrics are R@1. Bold indicates the best result for each benchmark.}
\label{tab:ens_retrieval}
\end{table*}

\begin{figure*}[t]
    \centering
\includegraphics[width=0.86\linewidth]{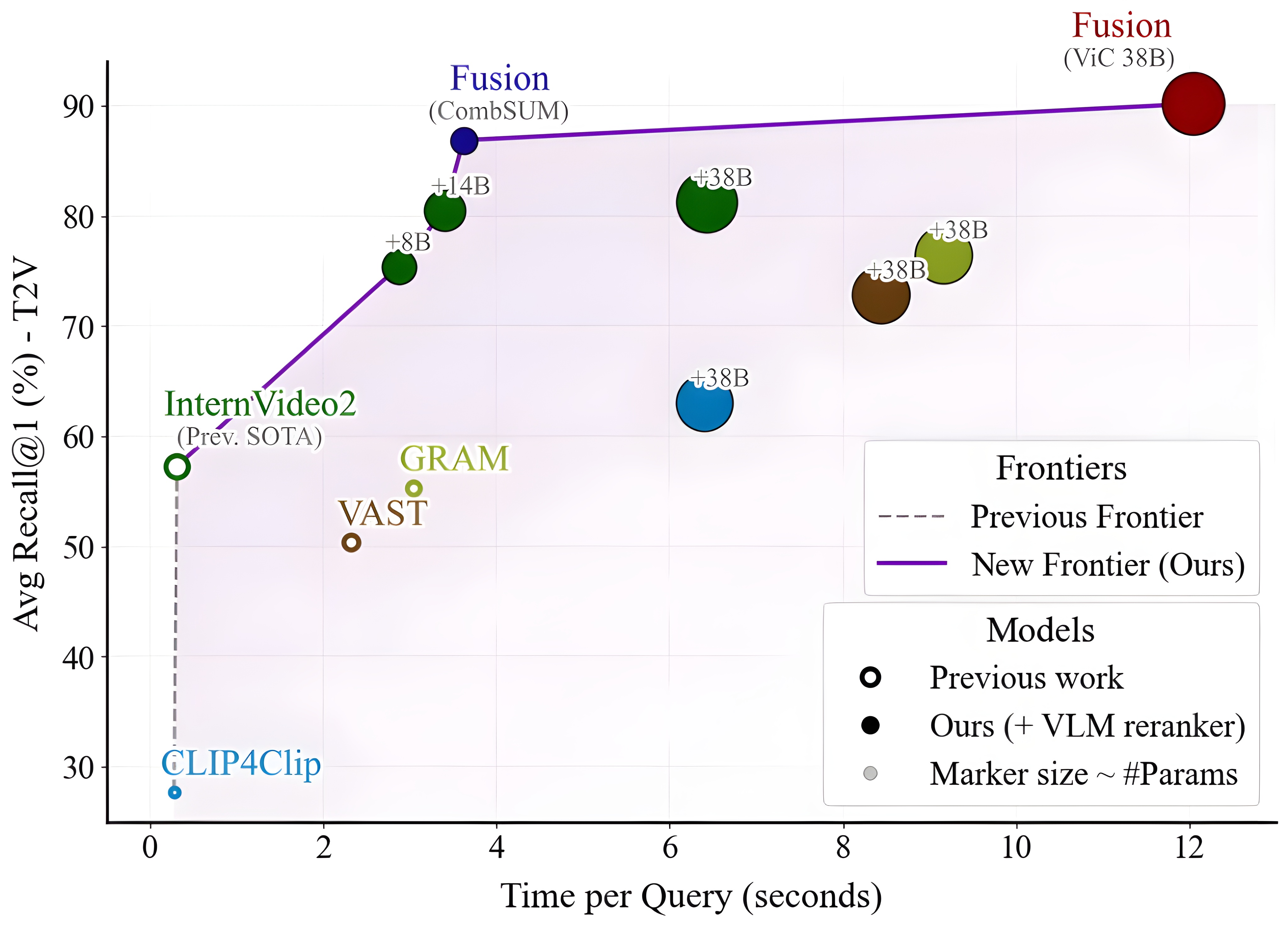}
    \caption{Efficiency vs. Performance Trade-off. Time per query vs. Avg Recall@1 for t2v retrieval over the benchmarks MSR-VTT, DiDeMo and ActivityNet in zero-shot settings. Marker size represents model parameters. The Pareto frontier highlights optimal trade-offs. Latency is measured on a single NVIDIA A100 80GB GPU, averaged over 50 queries for a 1k video retrieval task.}
    \label{fig:placeholder}
\end{figure*}

\section{Experiments}
\label{sec:results}

\subsection{Benchmarks and Protocol}
Evaluation is conducted on the MSR-VTT~\cite{xu2016msrvtt}, DiDeMo~\cite{anne2017didemo}, ActivityNet Captions~\cite{krishna2017activitynet}, and VATEX~\cite{wang2019vatex} benchmarks, following the standard retrieval protocols established in prior work. Notably, only MSR-VTT and VATEX provide subtitles, which are incorporated into the S-Grid representation where applicable. On MSR-VTT, the standard 1k-A split is used. For DiDeMo, evaluation is performed at the video level by pooling the moment annotations into a single retrieval target per video. ActivityNet Captions is evaluated using the official validation split for retrieval. For VATEX, the community 1.5k test subset is adopted. Out of the intended 1,500 videos from prior work, only 1,252 were successfully recovered due to the online unavailability of some videos. To ensure fair comparison, captions were re-indexed to this fixed subset, and all baselines and the proposed method were reproduced on the same 1,252 test videos. All evaluation items correspond to test-only instances, and the final video list is publicly released to facilitate reproducibility. Only msrvtt and vatex has subtitles.


\subsection{Implementation Details} The first-stage retrievers are CLIP4Clip\cite{luo2021clip4clip}, VAST\cite{chen2023vast}, GRAM \cite{cicchetti2024gramian}, and InternVideo2-6B \cite{wang2024internvideo2}. CLIP4Clip is a canonical CLIP-style video retriever. VAST provides omni-modality pretraining. GRAM is a strong global-regional baseline. InternVideo2-6B serves as the strongest recent baseline. Each model is reproduced or re-evaluated using official checkpoints and released evaluation configurations, and all retrievers are kept frozen during experimentation.
Tokenization, frame sampling, and text preprocessing strictly follow the original repository implementations to ensure consistency and reproducibility.

InternVL 3.5 38B \cite{wang2025internvl3} is employed as the main training-free VLM reranker. It consumes S-Grid inputs along with the video/text query and is used in a zero-shot setting without any dataset-specific fine-tuning. Unless otherwise noted, the same candidate counts are used for each comparison. The standard ensemble configuration fuses all backbones except VAST, as this combination yielded the highest performance on average.
A notable exception occurs in VATEX, where the ensemble includes only InternVideo2 and VAST, as these were the models successfully reproduced for this benchmark.


\subsection{Metrics and Hyperparameter}
Results are reported using Recall@1 (R@1), the proportion of queries for which the top-ranked result is correct, for both t2v and v2t directions. For t2v, the \textbf{ViC} framework receives $K=14$ candidate S-Grids per query, while for v2t, it receives $K=20$ candidate captions, unless stated otherwise. The default S-Grid size is $3 \times 3$ frames. For the Soft Voting baseline, similarity scores are min-max normalized per query (row) before aggregation with uniform weights. For \textbf{ViC} ensemble fuser ($M>1$), candidate lists are assembled by interleaving each retriever's list up to depth $k_{\max}$, preserving duplicates. The VLM output is parsed into a permutation, with the identity mapping used as a fallback in very rare cases. The resulting ranked list $\widehat{R}(q)$ may include duplicate candidates; however, only the highest-ranked instance of each is considered during evaluation, consistent with standard practice.

\begin{figure*}[t]
    \centering
    \includegraphics[width=1\linewidth]{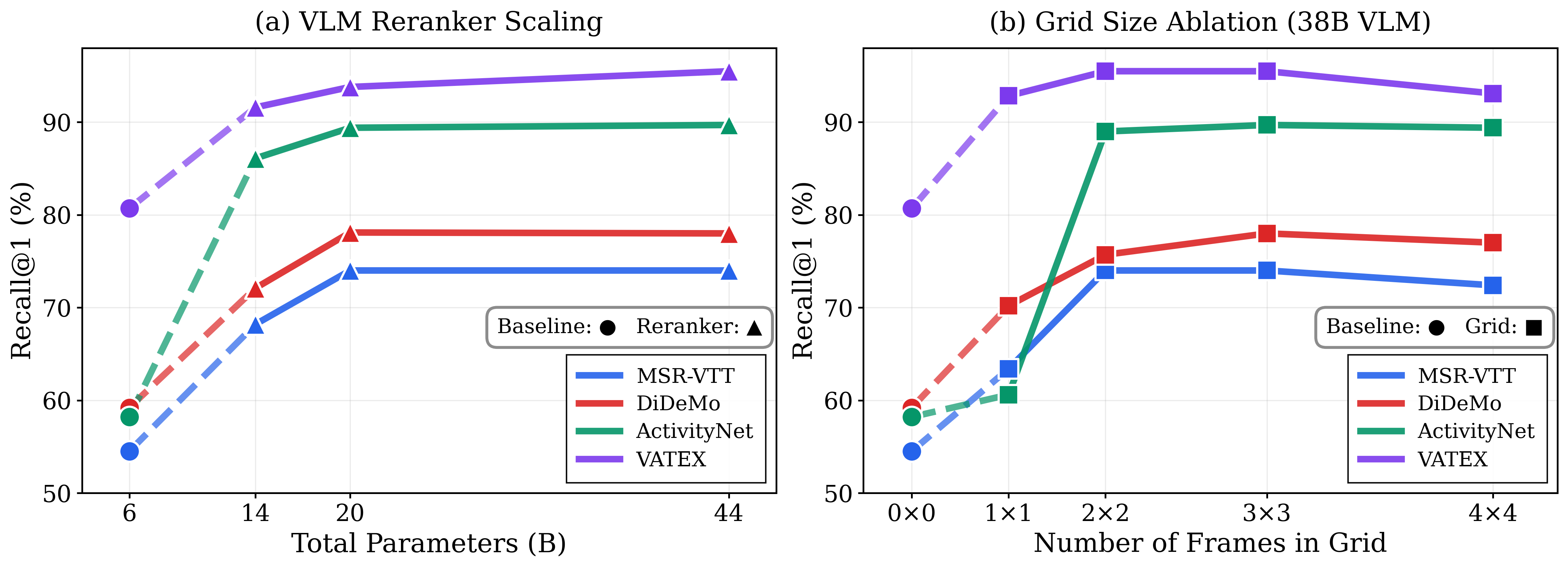}
    \caption{(a) Effect of reranker scale (InternVL 3.5, 3×3 grid) on t2v Recall@1.\;(b) Impact of grid size on t2v performance, using InternVideo2-6B and InternVL 3.5-38B.}
    \label{fig:gridsize_scale}
\end{figure*}

\begin{table*}[t]
\centering
\small
\setlength{\tabcolsep}{10pt}
\begin{tabular}{@{}l ccc ccc@{}}
\toprule
\multirow{2}{*}{\textbf{Reranker}} &
\multicolumn{3}{c}{\cellcolor{blue!15}\textbf{T2V — grids per query}} &
\multicolumn{3}{c}{\cellcolor{blue!15}\textbf{V2T — captions per query}} \\
\cmidrule(lr){2-4}\cmidrule(lr){5-7}
& \textbf{10} & \textbf{14} & \textbf{30} & \textbf{10} & \textbf{20} & \textbf{30} \\
\midrule
\rowcolor{gray!8}
& \textcolor{gray}{\scriptsize R@1/R@10} & \textcolor{gray}{\scriptsize R@1/R@10} & \textcolor{gray}{\scriptsize R@1/R@10}
& \textcolor{gray}{\scriptsize R@1/R@10} & \textcolor{gray}{\scriptsize R@1/R@10} & \textcolor{gray}{\scriptsize R@1/R@10} \\
InternVL 3.5 38B
& 73.8 / 82.7 & 74.0/ 83.8 & 71.3 / 84.5
& \textbf{75.8} / \textbf{85.3} & \textbf{74.1} / 89.0 & \textbf{70.0} / 90.8 \\
Qwen3-VL 30B (A3B)
& \textbf{76.5} / \textbf{82.7} & \textbf{77.0} / 84.7 & \textbf{77.0} / 86.5
& 59.5 / 85.3 & 55.2 / 84.2 & 51.8 / 85.4 \\
Gemma-3 27B IT
& 76.2 / 82.7 & 76.7 / \textbf{84.8} & 73.3 / \textbf{88.1}
& \textbf{75.8} / \textbf{85.3} & 71.2 / \textbf{90.5} & 69.3 / \textbf{91.5} \\
\bottomrule
\end{tabular}
\caption{Reranker type and context size in one view. Left: T2V vs.\ grids per query. Right: V2T vs.\ captions per query.}
\label{tab:unified_ctx}
\end{table*}

\section{Results}
\subsection{ViC as a Single-List Reranker (\texorpdfstring{$M=1$}{M=1})}
\label{ssec:results_single}

\textbf{ViC} is first evaluated in its simplest form as a single-list reranker ($M=1$), as defined in \S\ref{ssec:problem_setup}. In this setting, the VLM reranks the top-$K$ candidates from a single retriever, using only content evidence (S-Grids and subtitles) without any cross-list fusion metadata, as presented in Table~\ref{tab:retrieval}.

Applying \textbf{ViC} reranking to a single backbone yields substantial and consistent R@1 improvements across all datasets and models. For example, on MSR-VTT (t2v), \textbf{ViC} lifts the weakest backbone (CLIP4Clip) by 29.8 points (increases from 34.4 to 64.2) and the strongest (InternVideo2) by 21.4 points (increases from 54.5 to 75.9). On ActivityNet (t2v), the gains are even larger, adding 31.6 R@1 to InternVideo2 (increases from 58.2 to 89.8). On VATEX (v2t), \textbf{ViC} boosts VAST by 22.0 points (increases from 77.6 to 99.6), achieving near-saturation in R@1 performance.

These results demonstrate that, even without fusion, the VLM performs highly effective list-wise reasoning over S-Grid content evidence, providing a training-free mechanism to correct the coarse similarity biases of dual-encoder retrievers.
Moreover, a comparison between “Grid” (visuals only) and “S-Grid” (visuals and subtitles) configurations shows that incorporating textual evidence consistently enhances performance, confirming that the VLM effectively utilizes all available modalities during re-ranking.

\subsection{ViC as an Ensemble Fuser (\texorpdfstring{$M>1$}{M>1})}
\label{ssec:results_fusion}

The full \textbf{ViC} framework is evaluated as an ensemble fuser ($M>1$), utilizing both content evidence and retriever metadata (rank, multiplicity). A detailed comparison between \textbf{ViC} fusion and traditional fusion baselines (RRF, CombSUM, and CombMNZ) is summarized in Table~\ref{tab:ens_retrieval}

\textbf{ViC} consistently outperforms all traditional fusion methods across nearly all benchmarks. On MSR-VTT (t2v), \textbf{ViC} achieves 87.1 R@1, surpassing the best baseline (CombMNZ) by +1.8 points. On DiDeMo (t2v), the gain is most significant, where \textbf{ViC}'s 87.4 R@1 is +7.0 points higher than the next-best baseline (CombSUM). On VATEX (t2v), \textbf{ViC} reaches 97.5 R@1, once again setting the highest overall performance.

While RRF remains a strong competitor on ActivityNet, \textbf{ViC} demonstrates substantially greater stability across the other datasets, where RRF and other score-level fusion methods exhibit notable performance fluctuations.

Furthermore, Table~\ref{tab:ens_retrieval} includes a \textbf{ViC} (No Duplicates) ablation. This variant deduplicates the candidate sequence $C(q)$ before passing it to the VLM, thus removing the multiplicity metadata. The resulting performance drop (such as 87.1 to 84.2 on MSR-VTT t2v) confirms that the VLM actively uses cross-list consensus as a strong relevance signal.

Finally, comparing the \textbf{ViC} fusion result (87.1 on MSR-VTT, Table~\ref{tab:ens_retrieval}) with the best single-backbone re-ranking result (75.9 on MSR-VTT, Table~\ref{tab:retrieval}) highlights the additive advantage of fusion. 
Re-ranking a single model (\textbf{ViC}, $M = 1$) yields a +21.4 point improvement, while incorporating fusion (\textbf{ViC}, $M > 1$) contributes an additional +11.2 points, underscoring the complementary strengths of the two components within the \textbf{ViC} framework. 
This consistent, state-of-the-art performance across all benchmarks is visualized in Figure~\ref{fig:radar}. 

Moreover, Figure~\ref{fig:placeholder} contextualizes these performance gains against their inference cost. It clearly shows that \textbf{ViC}'s reranking and fusion methods establish a new, dominant Pareto frontier. While the original retrievers, such as InternVideo2, are fast, their performance is limited, clustering at the bottom-left. In contrast, \textbf{ViC} provides a massive leap in average R@1, pushing the SOTA from ~57\% to ~90\%. This gain comes at the expected latency cost of a second-stage reranker. However, the frontier itself shows promising scaling: the 8B and 14B models already achieve strong results, suggesting that the barrier to high performance is low and that future work on lightweight, fine-tuned rerankers could offer an even better performance-cost balance.

\subsection{Ablation Studies}

\subsubsection{Grid size}
As \textbf{ViC} relies on content-derived evidence, the S-Grid constitutes the key visual representation driving its performance. Figure~\ref{fig:gridsize_scale} (b) studies $1\times1$ to $4\times4$ grids. $2\times2$ and $3\times3$ are the sweet spots. $1\times1$ undercovers the video. $4\times4$ begins to compress each frame too aggressively and can introduce redundant visual tokens. This trend holds across the benchmarks that have been tested. Small grids are well matched to the evaluated datasets: MSR-VTT uses 10-30 s clips, DiDeMo videos are about 25-30 s, and VATEX clips are around 10 s. ActivityNet Captions contains longer, untrimmed videos with average durations on the order of minutes, though, the reranker performs strongly.

\subsubsection{Reranker scale}
Scaling the VLM within the \textbf{ViC} framework from 8B to 38B parameters at a fixed $3{\times}3$ grid leads to a steady improvement in R@1, which eventually saturates with increasing model size, as shown in Figure~\ref{fig:gridsize_scale}(a).
Notably, even the 8B model achieves strong zero-shot re-ranking performance, whereas smaller models fail to produce consistent permutations.
This result identifies 8B as the minimum effective scale for zero-shot list-wise re-ranking in the proposed \textbf{ViC} pipeline. The strong performance of the 8B model, even without training, suggests that lightweight fine-tuning could be a very promising direction for developing highly efficient, much smaller rerankers.

\subsubsection{VLM Type and Context size}
Varying the number of candidates supplied to the VLM per query within the \textbf{ViC} framework significantly influences retrieval performance, as summarized in Table~\ref{tab:unified_ctx}. Preliminary analyses confirmed that R@30 is effectively saturated near 100\% across benchmarks, indicating that the correct item is almost always retrieved within the top 30 candidates. However, the results indicate diminishing returns beyond a moderate context size.
For t2v retrieval, increasing $K$ from 10 to 14 yields higher R@1, but expanding to 30 causes R@1 to drop while providing only a negligible improvement in R@10. In practice, most VLMs fail to effectively utilize the additional coverage at $K = 30$, often exhibiting degraded discrimination accuracy due to overextended context.
Qwen3-VL, for example, performs strongly on t2v retrieval but deteriorates substantially on v2t when the context window increases.
Gemma-3 is an exception, maintaining stable performance at $K = 30$ and achieving the highest R@10 in both directions.
Nevertheless, InternVL 3.5 is employed in the main experiments owing to its consistent overall performance and the availability of multiple scale variants for systematic scaling analysis.
For v2t, an input size of $K = 20$ captions emerges as the most effective operating point, as performance plateaus or declines beyond this threshold.
These observations collectively highlight the practical limitations of current VLMs' effective context windows in list-wise relevance judgment tasks.

\subsection{Discussion and Conclusion}

The results make a compelling case for a new fusion paradigm: \textbf{ViC}.
Rather than relying on fixed formulas such as RRF or on trained fusers, \textbf{ViC} reconceptualizes fusion as a zero-shot, list-wise reasoning task performed by a VLM. This paradigm shift enables the model to adaptively balance retriever metadata, including rank and multiplicity, against content evidence on a per-query basis, leading to more context-aware and robust fusion behavior.

The practicality and effectiveness of the proposed framework are demonstrated in the challenging domain of video retrieval. By representing video as a compact S-Grid, we make it feasible for a VLM to process and re-rank an entire list of video candidates simultaneously. This representation maintains computational cost proportional to the number of candidates ($K$) rather than the raw video length, while still preserving temporal coverage and exploiting multimodal information from both visual and textual sources.
The resulting efficiency yields substantial R@1 improvements, enhancing user-perceived retrieval quality under fixed latency constraints and ultimately achieving state-of-the-art performance across benchmarks.

The limitations of the proposed approach can be categorized into those inherent to the \textbf{ViC} framework and those specific to its video-retrieval application.

The \textbf{ViC} framework itself introduces three main trade-offs. First, its inference cost is computationally expensive, replacing the near-zero arithmetic cost of RRF or CombSUM with a full, list-wise VLM forward pass. Figure~\ref{fig:placeholder} plots this trade-off, showing that \textbf{ViC} establishes a new Pareto frontier where it achieves significantly higher performance, albeit at a higher latency cost than traditional baselines. Second, the framework is strictly bounded by the VLM's context window, as our ablations show performance can degrade when $K$ increases. Finally, VLM reliability is a factor, as the framework's fidelity depends entirely on the VLM's instruction-following capabilities. Results might be influenced by positional bias or fail to parse the list format, as we observed with models smaller than 8B.

The video-retrieval application of \textbf{ViC} presents two additional limitations.
First, the method is inherently recall-bound: as with other two-stage retrieval systems, \textbf{ViC} cannot retrieve a relevant candidate if it is absent from the initial top-$K$ list produced by the first-stage retriever.
Second, while S-Grid serialization is computationally efficient, it remains inherently lossy, since uniform frame sampling from long, untrimmed videos may fail to capture short yet semantically important events that are essential for accurate query matching.

These limitations suggest clear directions for future work. At the framework level, promising approaches include prompt engineering and lightweight VLM fine-tuning to enable smaller, more efficient models to perform robustly.
At the application level, future research could investigate query-aware or adaptive keyframe selection mechanisms to generate more representative S-Grids within a fixed token budget.

{
    \bibliographystyle{unsrt}   
    \bibliography{main}
}

\clearpage